\documentclass[pdflatex,sn-mathphys-num,hyperfootnotes=true]{sn-jnl}

\usepackage{graphicx}%
\usepackage{multirow}%
\usepackage{amsmath,amssymb,amsfonts}%
\usepackage{amsthm}%
\usepackage{mathrsfs}%
\usepackage[title]{appendix}%
\usepackage{xcolor}%
\usepackage{textcomp}%
\usepackage{manyfoot}%
\usepackage{booktabs}%
\usepackage{algorithm}%
\usepackage{algorithmicx}%
\usepackage{algpseudocode}%
\usepackage{listings}%
\usepackage{acro}
\usepackage{placeins}
\usepackage{tablefootnote}
\usepackage[capitalise,noabbrev]{cleveref}
\usepackage{tabularx}
\usepackage{array}
\usepackage{booktabs} 
\usepackage{siunitx}
\usepackage{adjustbox}
\usepackage{subcaption}
\usepackage[normalem]{ulem}

\newcolumntype{C}{>{\centering\arraybackslash}X}

\DeclareAcronym{lm}{
  short=LM,
  long=Levenberg-Marquardt,
}
\DeclareAcronym{ar}{
  short=AR,
  long=average recall,
}
\DeclareAcronym{ap}{
  short=AP,
  long=average precision,
}
\DeclareAcronym{iou}{
  short=IoU,
  long=intersection-over-union,
}
\DeclareAcronym{svd}{
  short=SVD,
  long=singular value decomposition,
}
\DeclareAcronym{se}{
  short=SE,
  long=SurfEmb,
}
\DeclareAcronym{ese}{
  short=ESE,
  long=EpiSurfEmb,
}
\DeclareAcronym{fp}{
  short=FP,
  long=FoundationPose,
}
\DeclareAcronym{mvfp}{
  short=MVFP,
  long=Multi-view-FoundationPose,
}

\theoremstyle{thmstyleone}%
\theoremstyle{thmstyletwo}%

\theoremstyle{thmstylethree}%
\newcommand{\rev}[2][]{#2}
\newcommand\revTwo[1]{#1}

\begin{document}

\title{Training-free Detection and 6D Pose Estimation of Unseen Surgical Instruments}


\author*[1,2]{\fnm{Jonas} \sur{Hein}}\email{jonas.hein@inf.ethz.ch}

\author[1]{\fnm{Lilian} \sur{Calvet}}

\author[1]{\fnm{Matthias} \sur{Seibold}}

\author[2]{\fnm{Siyu} \sur{Tang}}

\author[2]{\fnm{Marc} \sur{Pollefeys}}

\author[1]{\fnm{Philipp} \sur{Fürnstahl}}

\affil[1]{\orgname{University Hospital Balgrist, University of Zurich}, \orgaddress{\country{Switzerland}}}

\affil[2]{\orgname{ETH Zurich}, \orgaddress{\country{Switzerland}}}


\abstract{ %
\textbf{Purpose:}
Accurate detection and 6D pose estimation of surgical instruments are crucial for many computer-assisted interventions. 
However, supervised methods lack flexibility for new or unseen tools and require extensive annotated data. 
This work introduces a training-free pipeline for accurate multi-view 6D pose estimation of unseen surgical instruments, which only requires a textured CAD model as prior knowledge.

\textbf{Methods:} 
Our pipeline consists of two main stages. 
First, for detection, we generate object mask proposals in each view and score their similarity to rendered templates using a pre-trained feature extractor. 
Detections are matched across views, triangulated into 3D instance candidates, and filtered using multi-view geometric consistency. 
Second, for pose estimation, a set of pose hypotheses is iteratively refined and scored using feature-metric scores with cross-view attention.
The best hypothesis undergoes a final refinement using a novel multi-view, occlusion-aware contour registration, which minimizes reprojection errors of unoccluded contour points.

\textbf{Results:}
The proposed method was rigorously evaluated on real-world surgical data from the MVPSP dataset. The method achieves millimeter-accurate pose estimates that are on par with supervised methods under controlled conditions, while maintaining full generalization to unseen instruments. 
These results demonstrate the feasibility of training-free, marker-less detection and tracking in surgical scenes, and highlight the unique challenges in surgical environments.

\textbf{Conclusion:} 
We present a novel and flexible pipeline that effectively combines state-of-the-art foundational models, multi-view geometry, and contour-based refinement for high-accuracy 6D pose estimation of surgical instruments without any task-specific training. 
This approach enables robust instrument tracking and scene understanding in dynamic clinical environments.

}


\maketitle

\section{Introduction}\label{introduction}

 

Accurate and reliable 6D pose estimation of surgical instruments is a fundamental task for computer-assisted interventions. 
It is crucial for a wide range of clinical applications, including surgical navigation \cite{molnar2020visual}, autonomous robotic actions \cite{xu2025surgripe,wu2025surgpose}, scene understanding \cite{lecuyer2020assisted}, and the development of digital twins for surgical training and education \cite{hein2024creating, ding2024digital}.
While long-available marker-based systems provide sub-millimeter accurate poses at high frame rates, their use is complicated by technical restrictions like calibration needs and line-of-sight issues, limiting their dissemination \cite{hartl2013worldwide}. 
Conversely, marker-less approaches based on deep learning methods offer significant potential for seamless integration into surgical workflows by reducing logistical and calibration overhead.

Traditionally, many approaches for this task have relied on supervised deep learning methods \cite{hein_next-generation_2025,xu2025surgripe,wu2025surgpose}. 
However, these methods suffer from significant practical limitations. 
They require time-consuming collection and laborious annotation of large, task-specific training datasets that have to be adapted to every new clinical use case. 
Surgical procedures rely on a diverse set of instruments, and variations in tool sets and manufacturers across institutions further amplify this issue.
Furthermore, these methods lack flexibility; introducing a new or modified instrument - or even one that was simply unseen during training, requires a complete and costly data collection and retraining cycle. 
These limitations hinder the development and maintenance of such methods in dynamic clinical environments.

The recent emergence of foundational models, trained on vast, multi-domain datasets, offers a promising alternative. 
Models like SAM2 \cite{ravi_sam_2024} and DINOv2 \cite{oquab2023dinov2} demonstrate remarkable zero-shot generalization abilities and can often be applied to novel domains, such as medical imaging, with little to no task-specific refinement. 
This potential opens the door for training-free methods that can adapt to new instruments on the fly.

Despite this promise, achieving the strict accuracy and robustness required for surgical applications remains a major challenge. 
This is particularly true for monocular pose estimation methods \cite{he_pvn3d_2020, wang_gdr-net_2021, haugaard2022surfemb, nguyen2024gigapose}, which are inherently limited by depth ambiguities and frequent occlusions common in complex surgical scenes. 
To overcome these limitations, we propose a novel pipeline that leverages the capabilities of foundational models within a new multi-view geometry framework. 
Our method is training-free and relies only on a reference CAD model of the instrument.
The main contributions of this work are threefold:
\begin{itemize}
    \item A complete, training-free pipeline for the detection and 6D pose estimation of surgical instruments from multiple views, based only on a reference CAD model.
    \item A multi-view matching and triangulation method to establish 3D instance candidates from 2D monocular detections, enabling multi-view pose estimation and refinement.
    \item A novel contour-based and occlusion-aware pose refinement algorithm that minimizes reprojection errors of unoccluded contour points.
\end{itemize}

\section{Related Works}\label{sec:relatedworks}

Our work is positioned at the intersection of classical 6D pose estimation, emerging training-free methods, and the specific domain of computer-assisted surgery.

\noindent\textbf{Supervised 6D Pose Estimation} 
The field of 6D object pose estimation has been dominated by supervised deep learning methods. 
These approaches learn a mapping from input pixels (RGB or RGB-D) to an object's 6D pose by training on large datasets of annotated image\rev{s}. 
Early methods like PoseCNN \cite{xiang_posecnn_2018} used an end-to-end approach that directly regresses rotation and translation parameters.
Subsequent works introduced intermediate representations such as 2D-3D correspondences \cite{wang_gdr-net_2021, haugaard2022surfemb} or keypoints \cite{he_pvn3d_2020}, enabling them to solve an interpretable geometric problem to compute the object's pose.
While supervised methods achieve state-of-the-art accuracy on benchmark datasets, their main limitation is the reliance on extensive, object-specific training data, limiting generalizability and rendering them impractical for real-world clinical use.

\noindent\textbf{Training-Free and Template-Based Methods}
To address the clinical need for generalizability, training-free methods leverage a 3D model or a set of reference images at test time without prior training on the specific object \cite{caraffa2024freeze, nguyen2024gigapose}. 
Previous approaches relied on matching local features \cite{lowe2004distinctive} or gradient-based templates \cite{hinterstoisser2012model}. 
More recently, CNOS \cite{nguyen_cnos_2023} demonstrated a powerful training-free pipeline by combining segmentation proposals with a feature-metric scoring, which serves as a key inspiration for our detection stage.
However, these methods often struggle with heavy occlusion or ambiguous monocular views. 
Our work extends this paradigm by integrating a robust multi-view aggregation and filtering pipeline, and by introducing a novel multi-view pose refinement stage.

\noindent\textbf{Foundational Models} 
The recent success of large-scale foundational models has introduced a new class of powerful, generalizable priors. 
DINOv2 \cite{oquab2023dinov2} provides dense, semantic features that are robust to viewpoint changes, while SAM2 \cite{ravi_sam_2024} offers exceptional zero-shot segmentation capabilities.
FoundationPose \cite{wen_foundationpose_2024} was a seminal work in this area, as the model showcased the emerging generalization capabilities that enabled object pose estimation across domains without any further refinement, by matching features from a rendered template to the query image.
Our work adopts the core components of these models, using SAM2 \cite{ravi_sam_2024} for mask proposals and DINOv2 for feature matching. 
We build upon the concepts of FoundationPose but adapt them for multi-view pose estimation and introduce a distinct contour-based refinement for higher accuracy.

\noindent\textbf{Surgical Instrument Pose Estimation}
In computer-assisted surgery, \revTwo{marker-less} instrument tracking is a long-standing challenge due to strong illumination, occlusions, and reflective, textureless surfaces \cite{doignon2005segmentation}. 
While several supervised methods have been proposed \cite{hein_next-generation_2025,xu2025surgripe}, they face the practical limitation in adapting to the wide variety of instruments used across different procedures and centers.
While multi-camera systems, such as those evaluated on the MVPSP dataset \cite{hein_next-generation_2025}, have shown promise in resolving ambiguities, they still rely on costly ground-truth annotations. 
Building on these advances, our work introduces the first practical solution to circumvent the need for ground-truth training data.
We propose a method that not only leverages multi-view geometry to overcome occlusions but also remains training-free, enabling a generalizable and scalable system for real-world surgical environments.

\section{Methods}\label{sec:methods}

\begin{figure}[t]
\centering
\hspace*{0pt}
\hfill
\includegraphics[width=\linewidth, trim={0 35mm 10mm 0}, clip, keepaspectratio]{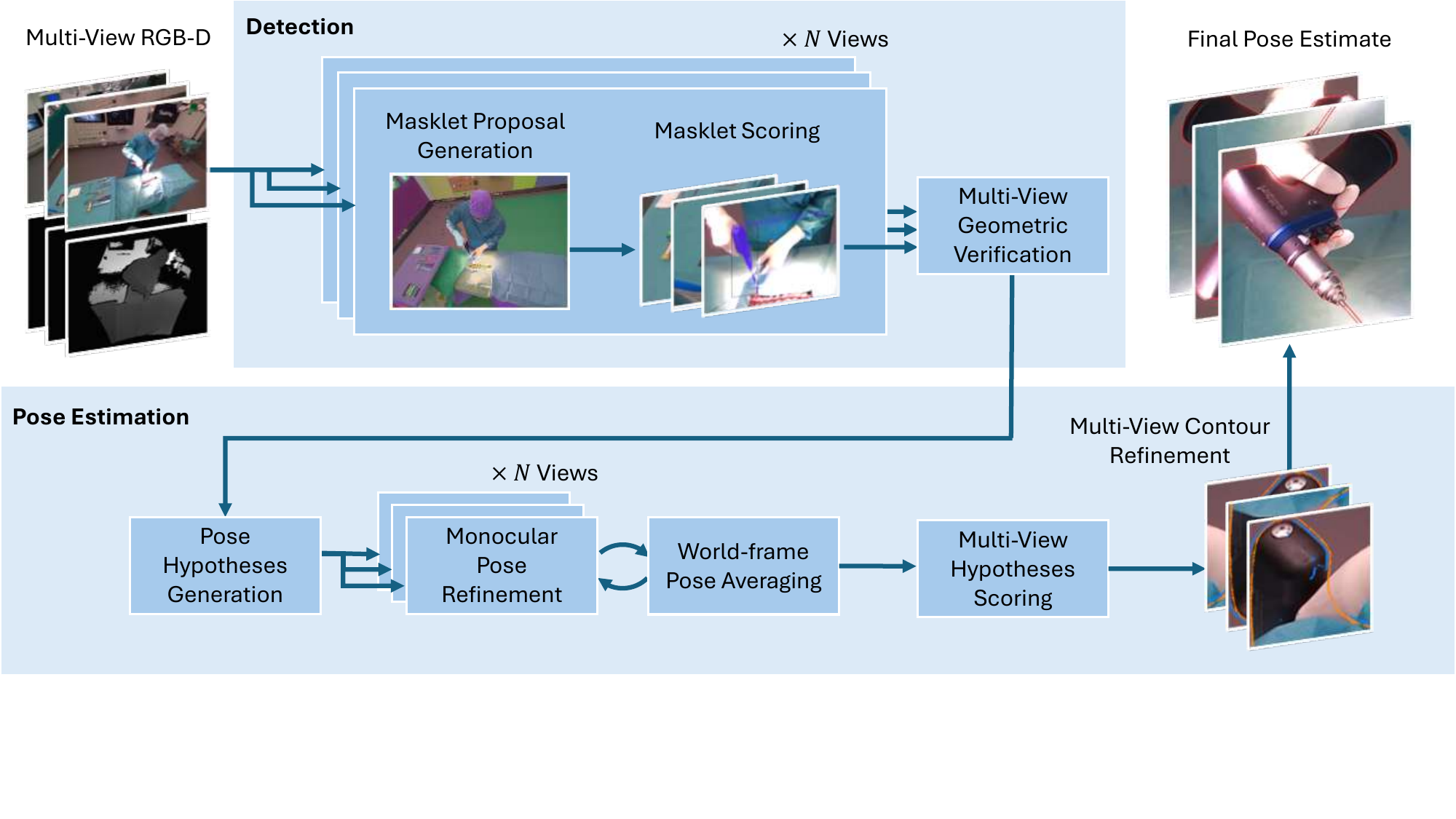}
\hfill
\hspace*{0pt}
\caption{Overview of our proposed pipeline. We first detect surgical instruments from multiple views using the detection stage outlined in \cref{sec:detection}. The 6D pose of all detected instances is then estimated using the pose estimation stage presented in \cref{sec:pose_estimation}.}
\label{fig:overview}
\end{figure}


Formally, the problem of detection and 6D pose estimation of novel surgical instruments in multiple views can be defined as follows:
Given a calibrated camera setup with $V$ views $\{v_1, ..., v_N\}$, known intrinsic $K_{v_i} \in \mathbb{R}^{3 \times 3}$ and extrinsic $\mathbf{T}_{v_i} \in SE(3)$ parameters, as well as a set of $O$ unseen target instrument classes $\{o_1, ..., o_M\}$, the goal is to detect all visible instances $\{q_1, ..., q_M\}$ and to estimate their poses $T_{q_i} \in SE(3)$.
Detected instrument instances in $v_i$ are represented as (modal or amodal) 2D bounding boxes $\hat{q_i} \in \mathbb{R}^4$ (detection task) or as binary masks $\hat{q_i} \in \{0,1\}^{H \times W}$ (segmentation task).
Our approach follows the established two-stage approach of solving the detection and pose estimation task separately, which we detail in the following sections.
\rev{In addition, we provide pseudo algorithms for both stages in the supplementary material.}

\subsection{Surgical Instrument Detection}\label{sec:detection}

Our detection stage is inspired by CNOS \cite{nguyen_cnos_2023} in that we first generate a set of class-agnostic object proposals, and score them against a set of pre-rendered templates of the test-time instruments.
In contrast to vanilla CNOS, we match masklets across multiple views $v_i$ and aggregate scores to enforce multi-view consistency.
This matching step is necessary as the multi-view pose estimation stage requires input patches from each view $v_i$ that correspond to the same instance, and furthermore enables a cross-view score aggregation. 

We use the foundational segmentation model SAM2 \cite{ravi_sam_2024} for each view $v_i$ to generate a set of $N$ class-agnostic object proposals $\{m_1, ..., m_N\}$.
Each masklet is scored independently based on its similarity to the test-time instruments, by comparing it to a pre-computed set of rendered instrument templates.
For each instrument $o_i$, we pre-compute a set of $K$ rendered template views by sampling poses from a standard viewsphere around the instrument's CAD model, resulting in a set of reference templates $\{t_1, ..., t_{O \cdot K}\}$ containing all instruments.
The template similarity scores $S_t(m_i, t_j)$ are computed as the cosine similarity between their respective DINOv2 features \cite{oquab2023dinov2}.
Following \cite{nguyen_cnos_2023}, we aggregate the scores for each instrument as the mean of its top-5 template scores, denoted as $S_c(m_i, o_i)$.

To aggregate the scores across the available views, we first triangulate 3D instance center candidates from pairs of masklets.
For each masklet pair $(m_i, m_j)$ from distinct views $v_i \neq v_j$, we triangulate the intersection point of the 3D camera rays that pass through the center of the masklets bounding boxes, more specifically the 3D center point $p_{m_i,m_j}$ of the shortest connecting segment between the two rays.
For each intersection point $p_i$, we compute its per-class scores as the product of the masklet class scores
$S_p(p_{(m_i,m_j)}, o_k) = S_c(m_i, o_k)\cdot S_c(m_j, o_k)$ and assign it the class with the highest score $C(p_i) = \text{argmax}_k S_p(p_i, o_k)$.

Next, we cluster the resulting point cloud $P = \{p_1, ..., p_{N}\}$ to match masklets across views and map them to shared 3D instances. 
We utilize a greedy clustering approach that iteratively assigns the point $p_i$ with the highest score to an existing or a new cluster. 
A point $p_i$ with class assignment $C(p_i) = o_o$ is assigned to the spatially closest $c_j$ cluster if (1) the cluster has the same object class $o_o$ assigned, (2) the euclidean distance between the point $p_i$ and the cluster center $c_j$ is smaller or equal to the object diameter $d_{o_o}$, and (3) after adding $p_i$, the cluster is supported by at most one masklet per view. 
These three criteria ensure that (1) clusters have a unique object class, (2) the diameter of the cluster does not exceed the physical size of the object, and (3) each cluster is supported by at most one masklet from each view.
If no cluster fulfills these criteria, $p_i$ is defined as a new cluster center.
From the clustered point cloud we then extract 3D instances $q_i = (c_i, \{m_j, ..., m_k\})$ comprising sets of masklets from unique views and approximate 3D instance center points $c_i$.

\subsection{Pose Estimation}\label{sec:pose_estimation}

For each detected object, the pose estimation stage follows a two-step, coarse-to-fine approach. 

\noindent\textbf{Coarse Pose Estimation}
\rev{Our \ac{mvfp} model builds} on top of FoundationPose \cite{wen_foundationpose_2024} to estimate a coarse pose hypothesis for each detected object instance $q_i$, while taking all available views into account.
Similar to the original method, we refine and score a set of pose hypotheses; however, in our formulation, the refinement is performed in parallel for each view. 
The resulting per-view estimates are then aggregated to ensure geometric consistency across views.
In the scoring step, we extend FoundationPose's attention-based scoring to operate across all views.
These extensions are implemented without requiring to modify the architecture of the underlying transformers, enabling us to retain the pre-trained models for a training-free pipeline.

We first generate a set of coarse pose hypotheses by uniformly sampling $N$ rotations from an icosphere centered at the estimated 3D instance center point $c_i$, expressed in a world coordinate frame. 
Next, each pose hypothesis is iteratively refined using FoundationPose's refinement network, which renders the object under the given pose hypothesis and estimates a rotation update $\Delta R$ and a translation update $\Delta t$ based on the comparison of the rendered object with the RGB input patch.
As the refinement network operates in the camera coordinate frame, we transform each pose hypothesis from the world to the corresponding view frame using known extrinsics $\mathbf{T}_{v_j}$, and transform it back to the world frame after refinement.
We enforce the consistency of all refined pose hypothesis by averaging the per-view rotations (using \acl{svd}) and translations in the world coordinate frame.
This step effectively shares information across views and pushes the pose hypotheses to converge towards a pose that explains all views.

The best pose hypothesis is selected by scoring each pose hypotheses relative to one another, using the same two-stage comparison as FoundationPose.
The pose ranking encoder computes $N \cdot V$ feature embeddings $\mathbf{F} \in \mathbb{R}^{512}$ that describe the visual alignment between the input patch and the rendered object hypothesis.
FoundationPose treats these features as a sequence, on top of which multi-head self-attention is applied.
We leverage this design and extend the sequence to include features from all views, stacked into a sequence $[\mathbf{F}_1, ..., \mathbf{F}_{N \cdot V}]$.
Last, the attended features are projected to scalar scores $S \in \mathbb{R}^{N \times V}$ and averaged across all views.
The pose with the highest average score across all views is selected as the final estimate $\mathbf{T}_{\text{coarse}}$.

\noindent\textbf{Multi-View Contour Refinement}
Following common practice, we apply a pose refinement stage to further increase the accuracy of the pose hypothesis.
The challenging illumination in surgeries combined with the reflective and largely textureless instrument surfaces lead to great variations in the visual appearance and limit the reliability of surface-based cues.
In contrast, instrument boundaries can be robustly extracted.
These observations motivate our contour-based refinement step, which jointly minimizes the reprojection error of the instrument contours in all views. 

We extract the set of 3D contour points $\mathbf{X}_i^\text{mask}$ from the rendered object mask under the pose hypothesis.
Similarly, we reuse the matched masklets ${m_i}$ from our object detection stage, and extract the set of 2D contour points $\mathbf{X}_i^{rgb}$ on the boundary of each masklet.
Here, a key challenge is that these contours may contain points from both the instrument and any occluding objects. 
To filter these, we compute local surface gradients from the instrument's depth map. 
We compute the 2D outward normal $n^\text{mask}$ for each rendered contour point $x_m \in \mathbf{X}_i^\text{mask}$, and the 3D surface normal $n^\text{surf}$ from depth, project it to $n^\text{proj}$, and discard the point if $n^\text{mask} \cdot n^\text{proj} \leq 0$, yielding a set of unoccluded contour points $\mathbf{X}^\text{visib} \subseteq \mathbf{X}^\text{rgb}$.
This effectively removes contour points generated by an occluding edge rather than the instrument itself.

Finally, \revTwo{we} refine the pose $\mathbf{T}$ using a \ac{lm} optimization. 
The optimization jointly minimizes a robust $L_1$ reprojection error between the set of unoccluded contour pixels $\mathbf{X}^\text{visib}$ and the closest projected 3D model contour points $\Pi_v(\mathbf{T}, \mathbf{X}_v^\text{mask})$ across all views $v$:
$$
\mathbf{T}_{\text{final}} = \arg \min_{\mathbf{T}} \sum_{v \in V} \sum_{x_v \in X_v^\text{visib}} \rev{\min_{x_m \in \mathbf{X}_v^\text{mask}}} \left\| x_v - \Pi_v(\mathbf{T}, x_m) \right\|_1
$$
\noindent where $\Pi_v$ is the projection operation into view $v$ based on the camera's known intrinsics and extrinsics. 
In our experiments, we recompute and re-match the rendered contour points every 10 optimization steps. We run the optimization for 200 iterations, which we empirically found to be sufficient for convergence.

\begin{table*}[t]
\centering
\begin{adjustbox}{width=1\textwidth}
\begin{tabularx}{1.2\textwidth}{lCCCCCC} 
\toprule
\multicolumn{1}{c}{\multirow{2}{*}{Model}} & \multicolumn{2}{c}{Synthetic} & \multicolumn{2}{c}{MVPSP Wetlab} & \multicolumn{2}{c}{MVPSP OR-X} \\
\cmidrule(lr){2-3} \cmidrule(lr){4-5} \cmidrule(lr){6-7}
\multicolumn{1}{c}{} & Mask AP & BBox AP & Mask AP & BBox AP & Mask AP & BBox AP \\ 
\midrule
SAM2+CNOS & 0.16 & 0.13 & 0.14 & \textbf{0.14} & 0.34 & 0.16 \\
\rev{+ 2-view verification} & 0.06 & 0.04 & 0.14 & 0.12 & 0.35 & \textbf{0.18} \\
\rev{SAM2+Oracle} & \textbf{0.54} & \textbf{0.57} & 0.23 & \textbf{0.14} & 0.43 & 0.17 \\
\rev{+ 2-view verification} & 0.47 & 0.46 & \textbf{0.24} & 0.13 & \textbf{0.48} & \textbf{0.18} \\
\midrule
SAM2+CNOS & 0.16 & 0.12 & 0.13 & 0.10 & 0.21 & 0.10 \\
\rev{+ 5-view verification} & 0.16 & 0.14 & 0.15 & 0.12 & 0.21 & 0.10 \\
\rev{SAM2+Oracle} & 0.54 & 0.57 & 0.24 & \textbf{0.15} & 0.34 & \textbf{0.14} \\
\rev{+ 5-view verification} & \textbf{0.61} & \textbf{0.60} & \textbf{0.25} & 0.14 & \textbf{0.35} & 0.12 \\
\bottomrule
\end{tabularx}
\end{adjustbox}
\caption{Comparison of our detection stage with and without multi-view verification on the MVPSP test sets and our synthetic dataset. We separately evaluate the performance on the subset of two cameras and the subset of five cameras and report the average precision (AP) \rev{over a range of \ac{iou} thresholds from \SIrange{0.5}{0.95}{} in steps of $0.05$}. The best average results are highlighted in bold font.}
\label{tab:results-detection}
\end{table*}

\section{Experiments}\label{sec:results}
We evaluate our approach on the MVPSP dataset \cite{hein_next-generation_2025}, which provides multi-view RGB-D captures of two surgical instruments in highly realistic ex-vivo surgeries.
We focus on two different camera configurations, namely a lightweight two-camera setup placed opposite of the head surgeon, and the heavier five-camera setup.

\begin{figure}[t]
\hspace*{0pt}
\hfill
\centering
\includegraphics[height=35mm, width=.49\linewidth, keepaspectratio]{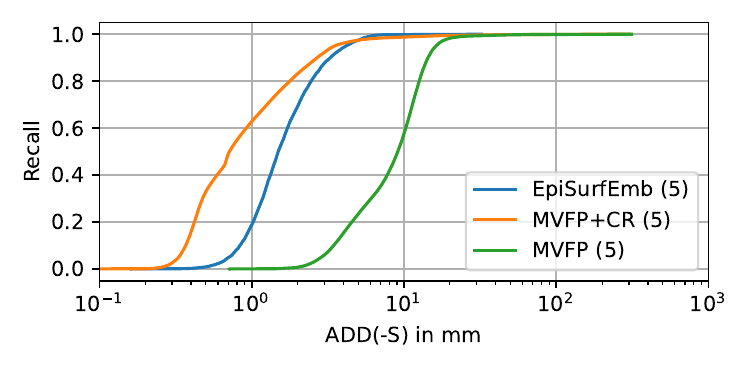}
\hfill
\includegraphics[height=35mm, width=.49\linewidth, keepaspectratio]{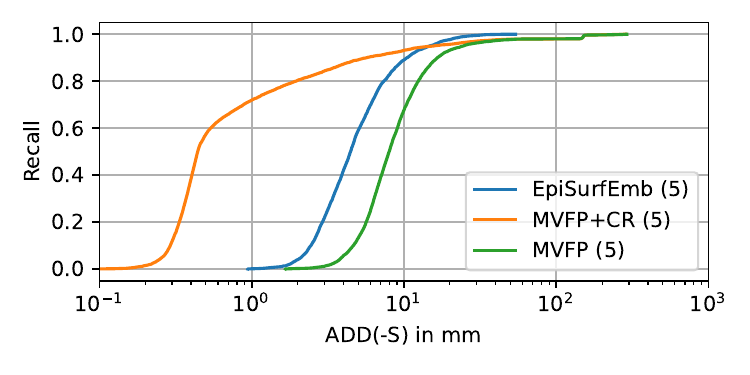}
\hfill
\hspace*{0pt}
\caption{\rev{Comparison of pose estimate error distributions from the supervised and training-free approaches. To enable a fair comparison to the supervised baseline, we use ground-truth detections in form of modal object masks as input to our pose estimation stage. All comparisons are done on 5 input views of the MVPSP wetlab test set (left) and the OR-X test set (right).}}
\label{fig:adds_error_distribution}
\end{figure}

We first evaluate the detection and pose estimation stages in isolation in \cref{tab:results-detection} and \cref{tab:results-pose-mvpsp}, respectively.
Following \cite{sundermeyer2023bop}, we report the \acf{ap} over a range of \ac{iou} thresholds from \SIrange{0.5}{0.95}{} for the detection task.
For the pose estimation stage, we use ground-truth detections in the form of modal object masks, enabling a comparison to the supervised methods evaluated in \cite{hein_next-generation_2025}. 
\rev{We report the ADD(-S) error \cite{hinterstoisser2012model}, the translation and rotation error as mean and standard deviation.
A comparison of the pose error distributions between the supervised and training-free approaches is shown in \cref{fig:adds_error_distribution}.
Finally, we evaluate our multi-view pose estimation stage with and without pose refinement on the detections from our detector stage \cref{tab:results-full-mvpsp}. 
Representative qualitative results are visualized in \cref{fig:qualitative_results,fig:failure-cases}.
Additional qualitative and quantitative results can be found in our supplementary material.}

\noindent\rev{\textbf{Ablation on Synthetic Images}}
\rev{Due to the limited number of instruments in the MVPSP datasets \cite{hein_next-generation_2025} we additionally evaluate our pipeline on a synthetic dataset including one representative instance from each of the 27  categories from the MedShapeNet database \cite{luijten_3d_2023, li_medshapenet_2025}.
Details on the generation of this dataset, as well as representative examples can be found in the supplementary material.
}

\noindent\rev{\textbf{Ablation with Masklet Scoring Oracle}}
\rev{The low detection scores of SAM2+CNOS prompt us to investigate the detection stage in more depth. 
In \cref{tab:results-detection,tab:results-pose-oracle} we evaluate an ablated version of our detection stage, where we replace the CNOS-inspired masklet scoring based on DINOv2 feature similarity with a \textit{scoring oracle}, which assigns each masklet a score that corresponds to their IoU with the ground-truth modal mask.
These results can be interpreted as an upper bound for the detection scores obtainable with SAM2-generated masklets, thus revealing both the missed potential in the masklet scoring step and the remaining potential of further improvements in the masklet proposal step.
}

\begin{table*}[t]
\centering
\begin{adjustbox}{width=1\textwidth}
\begin{tabularx}{1.6\textwidth}{lCCCCCC} 
\toprule
\multicolumn{1}{c}{\multirow{2}{*}{Model (Views)}} & \multicolumn{3}{c}{MVPSP Wetlab} & \multicolumn{3}{c}{MVPSP OR-X} \\
\cmidrule(lr){2-4} \cmidrule(lr){5-7}
\multicolumn{1}{c}{} & ADD(-S) & $\Delta t$ & $\Delta R$ & ADD(-S) & $\Delta t$ & $\Delta R$ \\ 
\midrule
SurfEmb* (1)    & 
$10.12 \pm 16.05$ & $12.09 \pm 18.63$ & $4.25 \pm 7.22$ & 
$40.52 \pm 58.06$ & $36.59 \pm 52.99$ & $7.92 \pm 14.71$ \\
EpiSurfEmb* (2) & 
$2.09 \pm 1.39$ & $\underline{1.97 \pm 2.15}$ & $\underline{2.46 \pm 3.58}$ & 
$8.06 \pm 6.31$ & $7.15 \pm 5.48$ & $3.00 \pm 2.56$ \\
EpiSurfEmb* (5) & 
$\underline{1.85 \pm 1.32}$ & $\mathbf{1.67 \pm 1.92}$ & $\mathbf{1.70 \pm 1.92}$ & 
$\underline{5.79 \pm 4.54}$ & $\underline{5.11 \pm 4.17}$ & $\mathbf{1.90 \pm 1.38}$ \\
\midrule
FoundationPose (1)        &
$61.20 \pm 89.95$ & $52.87 \pm 80.67$ & $44.88 \pm 54.48$ & 
$112.21 \pm 146.90$ & $90.47 \pm 131.86$ & $55.40 \pm 60.96$ \\
\rev{MVFP} (2)   & 
$31.68 \pm 72.16$ & $25.09 \pm 50.96$ & $22.23 \pm 41.35$ &
$19.96 \pm 40.34$ & $13.91 \pm 33.04$ & $10.19 \pm 21.88$ \\
\rev{\quad + CR} & 
$23.39 \pm 72.16$ & $16.05 \pm 52.48$ & $18.35 \pm 41.89$ & 
$13.14 \pm 37.41$ & $8.18 \pm 30.27$ & $7.00 \pm 21.91$ \\
\rev{MVFP} (5)   & 
$9.25 \pm 6.00$ & $11.25 \pm 6.01$ & $6.34 \pm 12.16$ &
$12.63 \pm 23.49$ & $9.99 \pm 15.60$ & $5.90 \pm 13.69$ \\
\rev{\quad + CR} & 
$\mathbf{1.50 \pm 5.22}$ & $2.09 \pm 5.15$ & $2.63 \pm 12.58$ & 
$\mathbf{5.44 \pm 24.48}$ & $\mathbf{3.75 \pm 16.68}$ & $\underline{2.74 \pm 14.53}$ \\ 
\bottomrule
\end{tabularx}
\end{adjustbox}
\caption{Results of our pose estimation stage on the MVPSP test sets. To enable a fair comparison to the supervised methods (*) evaluated in \cite{hein_next-generation_2025}, we use ground-truth detections in form of modal object masks as input. We report the ADD(-S) error, translation and rotation errors in millimeters and degrees. Results are averaged over both instruments and take object symmetries into account. \rev{We compare our \ac{mvfp} model with and without contour-based refinement (CR). The number of input views for each method is indicated in parentheses.} The best average results are highlighted in bold font. \rev{The second-best results are underlined.}
}
\label{tab:results-pose-mvpsp}
\end{table*}

\begin{table*}[t]
\centering
\begin{adjustbox}{width=1\textwidth}
\begin{tabularx}{1.6\textwidth}{lCCCCCC} 
\toprule
\multicolumn{1}{c}{\multirow{2}{*}{Model (Views)}} & \multicolumn{3}{c}{MVPSP Wetlab} & \multicolumn{3}{c}{MVPSP OR-X} \\
\cmidrule(lr){2-4} \cmidrule(lr){5-7}
\multicolumn{1}{c}{} & ADD(-S) & $\Delta t$ & $\Delta R$ & ADD(-S) & $\Delta t$ & $\Delta R$ \\ 
\midrule
\rev{MVFP} (2)   & 
$94.02 \pm 150.14$ & $103.76 \pm 173.72$ & $50.69 \pm 56.90$ & 
$54.15 \pm 93.55$ & $44.81 \pm 95.25$ & $31.45 \pm 49.52$ \\
\rev{\quad + CR} & 
$99.32 \pm 152.94$ & $110.33 \pm 177.34$ & $52.13 \pm 56.89$ &
$52.56 \pm 94.86$ & $42.35 \pm 95.95$ & $30.19 \pm 50.81$ \\ 
\rev{MVFP} (5)   & 
$\mathbf{23.88 \pm 58.47}$ & $\mathbf{29.22 \pm 72.39}$ & $\mathbf{20.03 \pm 38.46}$ & 
$\underline{37.02 \pm 94.90}$ & $\underline{29.95 \pm 87.50}$ & $\underline{19.85 \pm 41.16}$ \\
\rev{\quad + CR} & 
$\underline{29.60 \pm 71.66}$ & $\underline{37.38 \pm 90.51}$ & $\underline{22.08 \pm 40.61}$ &
$\mathbf{33.26 \pm 99.05}$ & $\mathbf{26.60 \pm 90.80}$ & $\mathbf{17.34 \pm 42.45}$ \\
\bottomrule
\end{tabularx}
\end{adjustbox}
\caption{Results of our pose estimation stage on the MVPSP test sets, based on the detections from our detection stage. We report the ADD(-S) error, translation and rotation errors in millimeters and degrees. Results are averaged over both instruments and take object symmetries into account. \rev{We compare our \ac{mvfp} model with and without contour-based refinement (CR). The number of input views for each method is indicated in parentheses.} The best average results are highlighted in bold font. \rev{The second-best results are underlined.}}
\label{tab:results-full-mvpsp}
\end{table*}

\begin{table*}[t]
\centering
\begin{adjustbox}{width=1\textwidth}
\begin{tabularx}{1.6\textwidth}{llCCCCCC} 
\toprule
\multicolumn{1}{c}{\multirow{2}{*}{Model (Views)}} & \multicolumn{1}{c}{\multirow{2}{*}{Scores}} & \multicolumn{3}{c}{Synthetic} & \multicolumn{3}{c}{MVPSP OR-X} \\
\cmidrule(lr){3-5} \cmidrule(lr){6-8}
\multicolumn{2}{c}{} & ADD(-S) & $\Delta t$ & $\Delta R$ & ADD(-S) & $\Delta t$ & $\Delta R$ \\ 
\midrule
MVFP (2) & CNOS & $85.96 \pm 93.64$ & $76.98 \pm 93.44$ & $97.62 \pm 63.83$ & $54.18 \pm 94.25$ & $44.89 \pm 94.91$ & $31.65 \pm 50.08$ \\
\quad + CR & CNOS & $107.41 \pm 119.93$ & $99.41 \pm 121.94$ & $101.18 \pm 60.16$ & $52.56 \pm 94.86$ & $42.35 \pm 95.95$ & $30.19 \pm 50.81$ \\
MVFP (2) & Oracle & $\underline{27.72 \pm 39.80}$ & $\underline{17.91 \pm 27.65}$ & $78.81 \pm 69.46$ & $20.39 \pm 29.72$ & $13.41 \pm 22.90$ & $12.51 \pm 22.56$ \\
\quad + CR & Oracle & $49.71 \pm 95.09$ & $40.73 \pm 93.39$ & $85.63 \pm 66.37$ & $17.72 \pm 31.55$ & $\underline{10.69 \pm 25.05}$ & $10.39 \pm 23.41$ \\
\midrule
MVFP (5) & CNOS & $37.67 \pm 57.52$ & $29.50 \pm 51.75$ & $73.16 \pm 70.33$ & $37.00 \pm 93.21$ & $29.91 \pm 86.63$ & $20.41 \pm 42.23$ \\
\quad + CR & CNOS & $66.30 \pm 104.85$ & $59.27 \pm 105.79$ & $80.41 \pm 65.11$ & $33.26 \pm 99.05$ & $26.60 \pm 90.80$ & $17.34 \pm 42.45$ \\
MVFP (5) & Oracle & $\mathbf{20.36 \pm 32.58}$ & $\mathbf{12.18 \pm 18.34}$ & $\mathbf{66.59 \pm 69.03}$ & $\underline{16.67 \pm 31.44}$ & $14.00 \pm 27.89$ & $\underline{9.84 \pm 23.45}$ \\
\quad + CR & Oracle & $33.80 \pm 65.84$ & $25.34 \pm 62.04$ & $\underline{73.97 \pm 66.08}$ & $\mathbf{12.07 \pm 32.27}$ & $\mathbf{9.72 \pm 28.62}$ & $\mathbf{7.15 \pm 24.14}$ \\
\bottomrule
\end{tabularx}
\end{adjustbox}
\caption{\rev{Comparison of the pose estimation accuracy when scoring masklets with CNOS versus a scoring oracle. We report the ADD(-S) error, translation and rotation errors in millimeters and degrees. Results are averaged over all instruments and take object symmetries into account. MVFP denotes our multi-view extension based on FoundationPose. The number of input views for each method is indicated in parentheses. The best average results are highlighted in bold font. The second-best results are underlined.}}
\label{tab:results-pose-oracle}
\end{table*}

\begin{figure}[t]
\hspace*{0pt}
\hfill
\centering
\includegraphics[height=35mm, width=\linewidth, keepaspectratio]{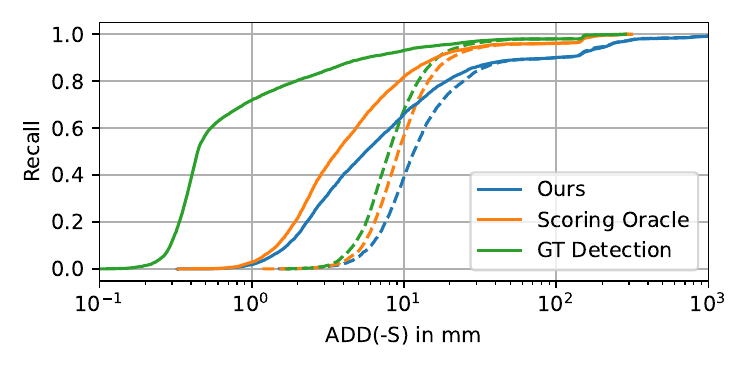}
\hfill
\hspace*{0pt}
\caption{\rev{Comparison of pose estimate error distributions depending on the instrument detections. We compare our detection stage against the SAM2 with a scoring oracle and the ground-truth detections with 5 input views on the MVPSP OR-X test set. Dashed lines indicate results of MVFP without refinement; solid lines of the same color show the results after contour-based refinement.}}
\label{fig:adds_error_distribution_oracle}
\end{figure}

\begin{figure}[t]
\hspace*{0pt}
\hfill
\begin{subfigure}{.3\textwidth}
\centering
\includegraphics[height=35mm, width=\linewidth, keepaspectratio]{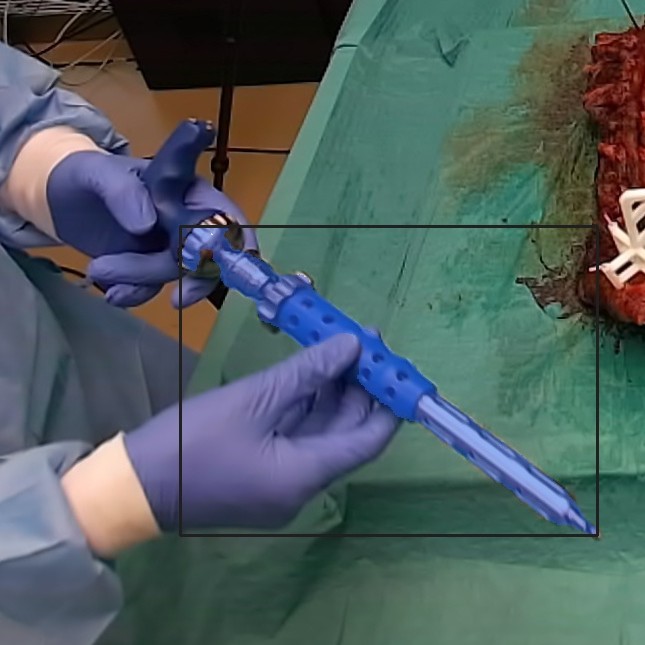}
\end{subfigure}%
\hfill
\begin{subfigure}{.3\textwidth}
\centering
\includegraphics[height=35mm, width=\linewidth, keepaspectratio]{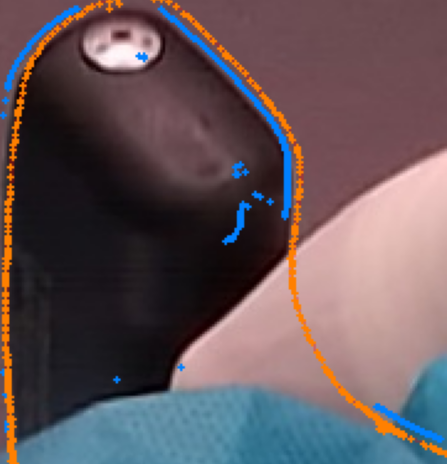}
\end{subfigure}%
\hfill
\begin{subfigure}{.3\textwidth}
\centering
\includegraphics[height=35mm, width=\linewidth, keepaspectratio]{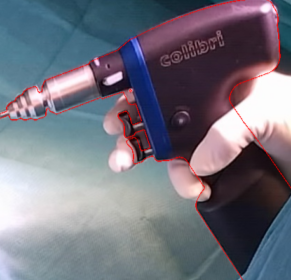}
\end{subfigure}%
\hfill
\hspace*{0pt}
\caption{
\rev{
Qualitative results from the detection and pose estimation stages. 
The left image shows an exemplary instrument detection of our 5-view model. 
The center image shows the extraction of unoccluded contour points, where the outline of the pose hypothesis is rendered in orange and the extracted contour points are highlighted in blue. Note the absence of contour points along the hand and the scrubs on the bottom of the image.
An exemplary result of the final pose estimate of our pipeline with five views and on ground-truth detections is displayed in the right image.
}}
\label{fig:qualitative_results}
\end{figure}

\begin{figure}[t]
\hspace*{0pt}
\hfill
\begin{subfigure}{.3\textwidth}
\centering
\includegraphics[height=35mm, width=\linewidth, keepaspectratio]{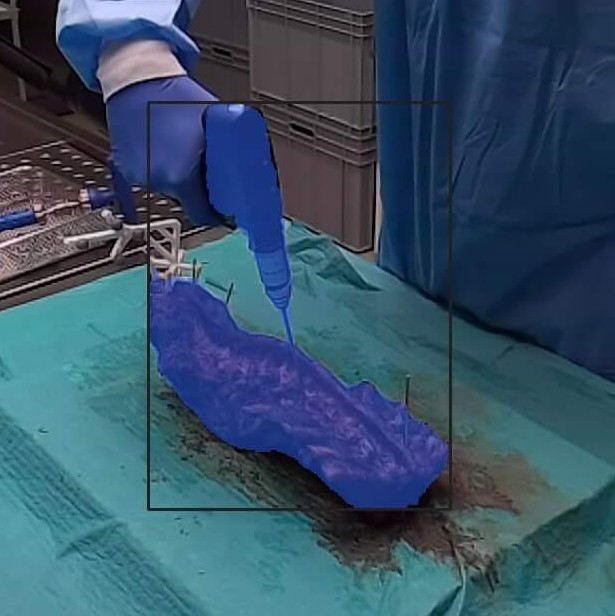}
\end{subfigure}%
\hfill
\begin{subfigure}{.3\textwidth}
\centering
\includegraphics[height=35mm, width=\linewidth, keepaspectratio]{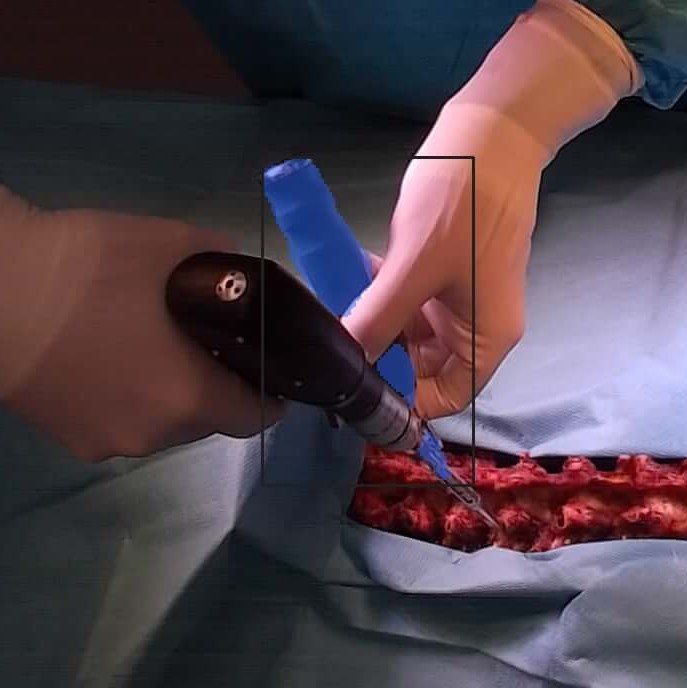}
\end{subfigure}%
\hfill
\begin{subfigure}{.3\textwidth}
\centering
\includegraphics[height=35mm, width=\linewidth, keepaspectratio]{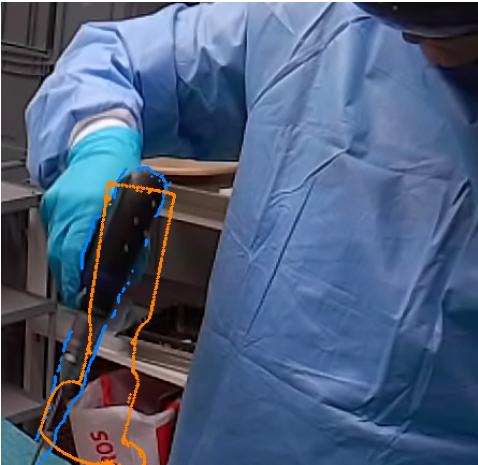}
\end{subfigure}%
\hfill
\hspace*{0pt}
\caption{\rev{Representative failure cases from the masklet generation (left), masklet classification (middle), and coarse pose estimation  step (right). We superimpose the detections with their predicted masklets and their bounding boxes in blue and black. For the pose estimation, we superimpose extracted contour points in blue and the estimated instrument pose as an orange outline. The coarse pose estimate is flipped along the instruments axis and unrecoverable for the refinement stage.}}
\label{fig:failure-cases}
\end{figure}

\section{Discussion}\label{sec:discussion}
Our work is the first to demonstrate the feasibility and potential of training-free, multi-view detection and 6D pose estimation of unseen surgical instruments. 
By combining the generalization capabilities of foundation models with the geometric consistency of a multi-view setup, the proposed approach bridges the gap between generic visual representations and precise 3D reasoning. 
The results highlight the challenges in the detection and pose estimation of unseen surgical instruments and the limitations in generalization of current methods.
Compared to existing benchmarks \cite{sundermeyer2023bop}, we observe a significantly larger performance gap between supervised and generalized methods, which can be attributed to the more challenging illumination in surgeries, as well as the prevalence of reflective and texture-less surface materials.

\noindent\textbf{Detection Stage}
The detection stage showed to be the main bottleneck in the end-to-end pipeline.
A closer analysis of the intermediate results from SAM2 and CNOS revealed the masklet classification as the main bottleneck\rev[, as shown in Figure 2]{}. 
While the instruments are often accurately segmented and the correct instrument masklets obtain consistently high scores, the model remains confused by other objects that share weak similarities with the target instrument, such as other instruments or camera tripod legs.
As these false positives are often consistent across views, they pass the multi-view verification.
Compared to the monocular baseline, the multi-view verification yields minor improvements in terms of \ac{ap}, which can be \rev{mainly} attributed to re-scored detections\rev[as the multi-view verification step does not re-estimate the segmentation masks or bounding boxes]{}. 
These results highlight the persisting challenge of a robust classification for unseen instruments due to the \rev{strong visual similarities of different surgical instruments and the} large domain gap between rendered instrument templates and their appearance in realistic surgical conditions.
Still, the multi-view verification step provides correspondences between detections in all views, which is necessary for the pose estimation stage, as well as a rough estimate of the instrument's 3D position.
\rev{Future extensions could utilize an anisotropic distance metric based on the instrument shapes to further improve the robustness of the clustering step.
Also, an optional fall-back to monocular estimates could improve the detection rates in challenging scenarios were an instance is detected only in a single view.}

\noindent\textbf{Pose Estimation Stage}
Our multi-view-consistent extension for FoundationPose yields a significant accuracy improvement over its monocular counterpart. 
Compared to the supervised baselines, the errors of FoundationPose are larger by a factor of $3-10$, depending on the metric and test set.
In this context, our multi-view extension bridges this gap considerably.
Still, the coarse pose estimation model has high potential for future enhancements. 
\rev{Comparing the 2-view and 5-view results of the supervised baseline and our training-free approach reveals the inherent advantage of supervised approaches compared to current foundation models.
While the supervised method yields comparable pose accuracy in both scenarios due to the strong shape and appearance priors, the more challenging generalization task of our training-free approach requires information from additional views to converge to an accurate pose estimate.
However, the gap between supervised and unsupervised methods has been closing in recent years \cite{sundermeyer2023bop}.
}
Thanks to the pipeline's modular design, more advanced methods can be readily integrated as they emerge.

The contour-based multi-view refinement provides a significant reduction of pose errors compared to the coarse estimates, as shown in \cref{tab:results-pose-mvpsp}.
In the five-view scenario with ground-truth detections, our training-free pose estimation stage yields pose errors on par with the best supervised alternatives.
However, the simple nearest-neighbor matching between extracted and rendered contour points significantly constrains the convergence basin and limits the robustness.
In the two-view scenarios and when using the detections from our detection stage, the initial pose estimates are not sufficiently accurate for the refinement to converge.
\rev{
Additional regularization based on the mask or depth consistency, or feature-metric correspondence matching, similar to existing 2D-2D correspondence based methods \cite{nguyen2024gigapose,Deng:CVPR2025}, could potentially improve the convergence characteristics in the future.}

\noindent\textbf{Limitations}
The presented results do not meet the requirements for our envisioned clinical applications yet, however, the modular design of our pipeline enables an individual enhancement of each component to benefit from future advances.
Still, this design also limits information sharing across steps, which could lead to accumulating errors.
Similarly, our approach benefits from reusing established monocular methods within a post-hoc multi-view aggregation framework, at the cost of hindering information flow due to the isolation of feature extraction and inference.
Multi-view feed-forward models inspired by recent works such as VGGT \cite{wang2025vggt} or MapAnything \cite{keetha2026mapanything} have significant potential to solve this limitation.

Although our experiments rely on a single dataset, it currently offers the most accurate multi-view RGB-D annotations available. 
Future work should extend this evaluation to broader and more diverse surgical scenarios.

\section{Conclusion}\label{sec:conclusion}
This work presents a feasibility study for a training-free detection and pose estimation of unseen surgical instruments.
The proposed pipeline combines the generalization capabilities of foundation models with the geometric consistency of a multi-view framework, requiring only a reference CAD model as prior knowledge. 
A multi-view, contour-based refinement step is introduced to improve pose accuracy, offering increased robustness under challenging illumination conditions.
Despite the remaining limitations, particularly regarding the use of current foundation models, we propose a promising direction toward geometry-driven solutions for computer-assisted surgery.

\backmatter

\section*{Declarations}

\noindent\textbf{Acknowledgements}
This work has been supported by the OR-X - a Swiss national research infrastructure for translational surgery - and associated funding by the University of Zurich, University Hospital Balgrist, and the InnoSuisse Flagship project PROFICIENCY (No. PFFS-21-19).

\noindent \textbf{Competing Interests} The authors have no relevant financial or non-financial interests to disclose.

\noindent \textbf{Ethics Approval}
Since this study did not involve human or animal subjects, ethics approval is not applicable.

\noindent \textbf{Informed Consent}
As this research did not involve human participants, informed consent is not applicable.

\noindent\textbf{Data availability}
All code and data related to this work will be made available online.

\bibliography{sn-bibliography}

\end{document}